\def\year{2021}\relax
\documentclass[letterpaper]{article} 
\usepackage{aaai21}  
\usepackage{times}  
\usepackage{helvet} 
\usepackage{courier}  
\usepackage[hyphens]{url}  
\usepackage{graphicx} 
\usepackage{natbib}  
\usepackage{caption} 
\usepackage{xcolor}
\title{The Science of Rejection: A Research Area for Human Computation}
\author{

    Burcu Sayin\textsuperscript{\rm 1}, 
    Jie Yang\textsuperscript{\rm 2}, 
    Andrea Passerini\textsuperscript{\rm 1},
    Fabio Casati\textsuperscript{\rm 3} \\
}
\affiliations{
    \textsuperscript{\rm 1}University of Trento\\ Via Calepina, 14, 38122 Trento TN, Italy \\
    \textsuperscript{\rm 2}Delft University of Technology\\ Mekelweg 5, 2628 CD Delft, Netherlands\\
    \textsuperscript{\rm 3}Servicenow\\ Santa Clara, CA, USA

}

\begin{document}

\maketitle

\begin{abstract}
We motivate why the science of learning to reject model predictions is central to ML, and why human computation has a lead role in this effort.
\end{abstract}


\section{Introduction}
For decades, the primary way to develop and assess machine learning (ML) models (and one of the primary avenues to publication) has been based on \textit{accuracy} metrics (e.g., precision, recall, F1, AUC). The need to beat leader-boards when publishing, with public datasets and rankings, to some extent exacerbated this focus on accuracy metrics, at least for classification models.
There is nothing bad in efforts to improve model accuracy: they should stay among the main goals of ML research. But we argue that one of the reasons for the disconnect between the amazing progresses of ML research (and the corresponding expectations of professionals in any field that are now sky high) and the limited adoption of ML in the enterprise is the focus on one aspect of the problem only, and on the fact that we have not paid enough attention to how and why models are used in practice, and to the aspects and metrics that are relevant to enterprises when they adopt and deploy a model. 

In this paper, we take an intentionally provocative stance and state that \textit{accuracy metrics are optional}, desirable properties of an ML model that are sometimes marginal with respect to other metrics rarely addressed in the literature, if at all. 
To this end, we start by taking a critical look at the use of ML models in typical enterprises scenarios, and from there abstract a simplified but general AI workflow followed in practice. We then present an analysis on how current metrics--- both the accuracy metrics and some more recent proposals---are misaligned with the value and cost induced from the workflow.
\textit{We come to conclusion that what we need is a new set of cost (loss) metrics as well as a science for learning when to reject the inferences done by an ML model - and correspondingly for identifying subsets of items where we can trust the model.} 

We then switch our discussion to the scientific progress related to those problems: where are we now and what can human computation do? We start by reviewing the ongoing work in ML and human computation, including the recent effort on data excellence and hybrid human-machine systems. We argue that human computation can play a lead role in providing methods for model rejection, yet discussions on this topic are largely missing. We discuss opportunities for research on metric definition, ML failure detection and characterization, and building the rejector, all involving crowds.

\section{AI Workflows and the Metrics that Matter}

\begin{figure}[]
\centering
\includegraphics[width=0.5\textwidth]{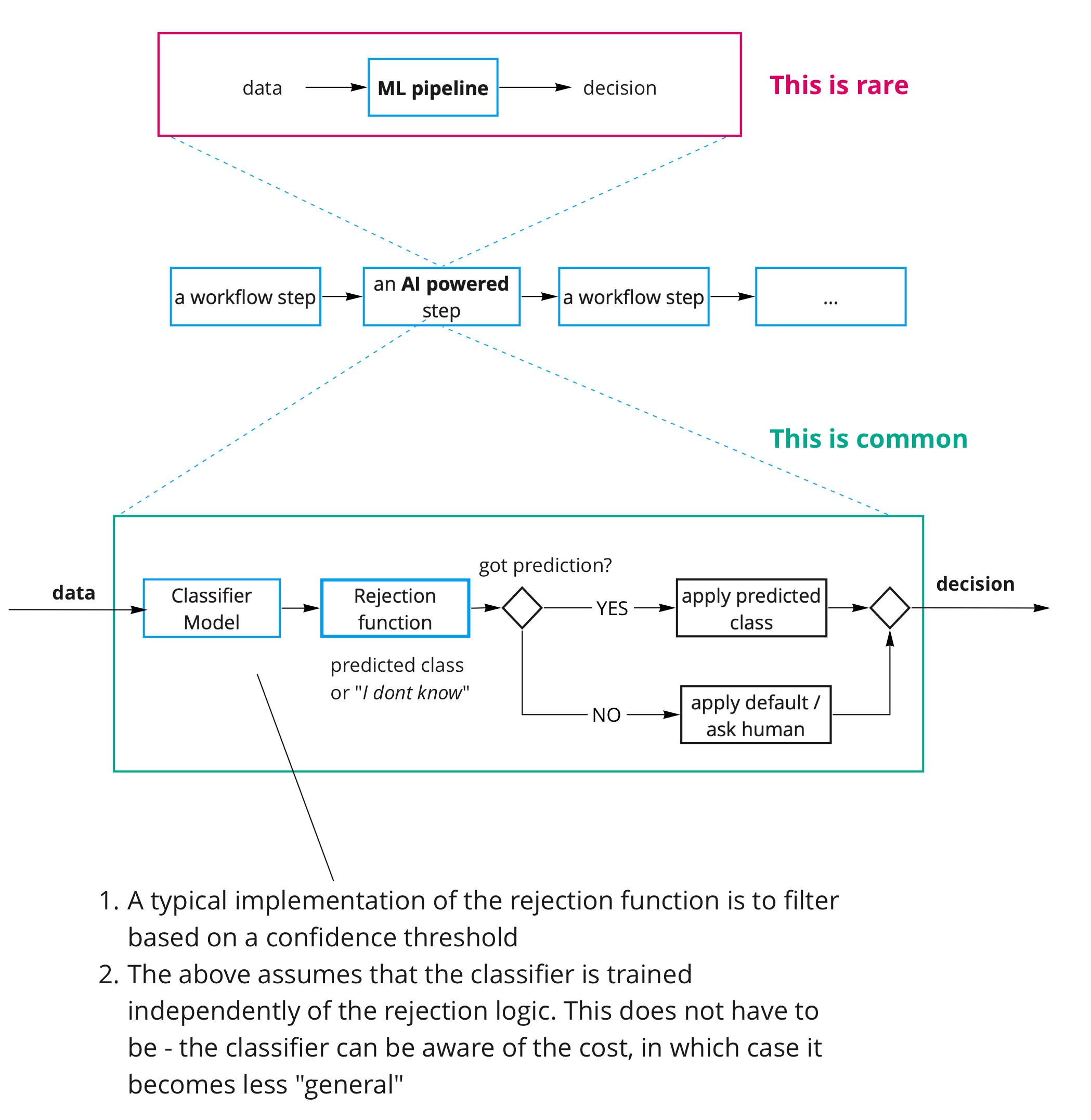}
\caption{Typical implementation of ML models into an enterprise workflow.}
\label{fig:wf}
\end{figure}

Figure~\ref{fig:wf} shows the typical way in which AI solutions are deployed in an end-to-end enterprise workflow. 
First, we either reuse or fine-tune a pre-trained model, or use it in combination with some task-specific model. Either way, we have some ML classifier $m$ that, given an input $i \in I$ (where $I$ is a possibly infinite set of items to classify), produces a predicted class and a confidence (or a distribution of predicted class with confidences).
Almost invariably today in any model deployment  there is then a filtering based on whether the confidence is greater than some threshold, and if so the prediction is applied, else a default path is followed - most often the very same path that was there before the introduction of AI in the process.
This is true from reading Xrays to detecting intents in a chatbot to automatically routing requests and on to nearly any use case we can imagine.

From this simple description we can draw a few observations:
First, the threshold and the system behavior depend on the ``cost'' of machine errors and its relation to the cost of a rejection and the value of a correct machine prediction. 
Let's refer to the value of a correct prediction as $V$, to the value (cost) of following the default flow as $C_d$ and to the cost of a wrong prediction as $C_w = K \cdot C_d$ (that is, we express $C_w$ in terms of how "bad" is an erroneous prediction compared to the default flow). 
Also, for simplicity, let's assume that $V = -C_d$ (this is both reasonable and to simplify the presentation, but none of the concepts below depends on this assumption), and let's normalize by taking $V = 1$, again for simplicity.
If the enterprise has a sense for the value of $K$, then the confidence threshold descends from that, and so does the expected utility we get from each execution of the workflow. The actual math and values are not relevant to the discussion, but simple math (omitted) shows that the optimal threshold $T$ is $T = \frac{K-1}{K+1}$, \textit{assuming the model is well calibrated}.
Similarly, we can show that the \textit{expected value} for each prediction with confidence $c$ is 

\begin{equation}
E[value] = -1 \cdot \rho_t  + (1 - \rho_t) \cdot (c(K+1) – K)
\end{equation}
where $\rho_t$ is the probability of a prediction confidence being below the selected threshold $t$.

The second observation is that if we have a well-calibrated model with \textit{arbitrarily bad accuracy} $\alpha$, we can still get value from it.
Having no model at all gives us a value of $-n$ to process $n$ items (remember $C_d=-1$). 
Having a really inaccurate (but well-calibrated!) model that, in the rare proportion of cases $hc$ where it has high confidence is correct, gives us a value of 
$ -n \cdot (1-hc) + n* hc \cdot$, which is higher.
In theory $hc$ \textit{can be arbitrarily small and we still obtain value}. 
In practice this is not the case as the decision to adopt AI comes with development, deployment, testing and management costs so there is some minimal value below which AI adoption does not make sense.


Notice that commonly adopted measures of calibration errors, such as the Expected Calibration Error (ECE) and its variation ~\cite{Nixon2019MeasuringCI} (eg, based on how we bin the samples) are not part of the above formulas and it is easy to show that they do not correspond to the metric we want to improve.

In this, it is important to point out that we are not just referring to far fetched corner cases.
The "problem" with commonly adopted metrics of calibration error is that, while they are valuable in helping us to get a sense of the model calibration as a whole and they are independent of any threshold $T$ or cost structure, that is also their limit. 
Indeed, when we apply a model as per the workflow above we only really care about calibration around the confidence threshold $T$, that either would take predictions we (incorrectly) reject above the threshold, or that take predictions we hazardously accept below the threshold, where they belong.
It is not uncommon for calibration techniques such as temperature scaling to show spectacular ECE results but if our threshold is 0.8, we really don't care about error in the 0.1-0.2 range, nor we care if a confidence is 0.999 or 0.85.

Other measures of calibration errors seem more meaningful in this regard: for example, we can measure the difference between \textit{expected value} as per the formula above (which does not depend on prediction correctness, only on confidence values) and the \textit{actual value} measured on a test set given the threshold $T$ based on actual correctness of predictions. 
If the model confidences represent probabilities and the model is calibrated, then the difference is only due to statistical sampling error, and as the test dataset increases in size it should go towards zero. 
If the model is not calibrated, then the difference has two sources: the error due to calibration and the error due to the fact that $T$ was set based on optimal calibration.
Another approach is to select the  threshold $T_v$ not based on the above formula but empirically based on a validation set, by picking the threshold that minimizes the cost over that dataset. 
In this case, the difference between the theoretical threshold $T$ (which assumes calibration) and $T_v$ grows as the calibration error grows.
These are just two examples of metrics around which we have not developed a theory yet but that seem promising in terms of defining a notion of calibration error that, when minimized, leads to a higher expected value.


Last but not least, the same discussion we had above for accuracy (that is: it's ok if we are accurate for an arbitrarily small subset of items) also applies to calibration! 
In principle, if we had a magic way to know that our model $m$ is well calibrated for a subset $I_c \subseteq I$ of items, and we knew, given an item $i$, how to tell if $i \in I$, then we would have a useful model, no matter what our cost structure is, no matter the accuracy, and no matter the overall calibration over $I$.

At some point the problem becomes recursive and we need to draw the line ( knowing $I_c$ is kind of the same thing as saying that we are confident about our confidence measures). But the main point remain: the better we are able to identify subset of items for which our model $m$ is calibrated - according to the metrics defined above - the lower is the cost for our deployment of $m$ in an AI workflow.

Now, because in this paper we assume that $m$ is given to us and that the decision to accept or reject examples is done downstream - as it often is in reality, this means that we need to equip AI practitioners with a "science of rejection" that helps build acceptance or rejection logic for each prediction - either by recalibrating a model and/or by identifying areas of $I$ where we can and cannot trust $m$.


\section{Where Are We Now?}

\smallskip
\noindent\textbf{Confidence Calibration} aims at making the model prediction confidence to be representative of the likelihood of the prediction to be correct. 
Typical methods smooth the training labels by converting a single hard label into a probability distribution using certain heuristics \cite{szegedy2016rethinking}, e.g., by reducing the probability of the label and amortizing the reduced probability over other labels. 
Such an approach has shown to be effective but again our concern is with the metric that the approaches optimize.
Furthermore, it has recently been shown that while label smoothing can prevent neural nets from becoming over-confident, it results in loss of information about resemblances between data instances \cite{muller2019does}. From the perspective of ML failures, label smoothing only deals with biases in the label and cannot deal with those in the feature space. 

\smallskip
\noindent\textbf{Adversarial Training} instead, can reduce such biases in the feature space by generating adversarial instances~\cite{szegedy2013intriguing}, also called out-of-distribution instances as they are not captured in the training data. The idea is developed driven by the observation that imperceptible differences in the processed data can lead to prediction failures. 
The approach however can lead to a skewed distribution of the generated instances that are similar to existing training instances. In particular, for certain features that are missing in the training data, it's unlikely that adversarial training can generate such data items. 


\smallskip
\noindent\textbf{Data Excellence} is a recent effort to enhance the quality of training data through human discovery of items that are challenging to ML models, especially the unknown unknowns (i.e., items on which the model makes high confidence errors)~\cite{attenberg2011beat}. Unlike machines that fully rely on knowledge explicitly encoded in predefined training data, humans excel at leveraging broad, tacit, and contextual knowledge in decision making and justification. Human computation has, therefore, emerged as a new, promising approach to detecting unknown unknowns. A seminal work by Attenberg \emph{et al.} propose to ask humans to gather publicly accessible instances that are potentially difficult for the model to handle~\cite{attenberg2011beat}. Lakkaraju \emph{et al.} introduce a data partitioning technique that first organises the data into multiple partitions based on feature similarity, and then uses an explore-exploit strategy to search for unknown unknown instances across these partitions~\cite{lakkaraju2017identifying}. An important finding in human computation studies reveals that unknown unknowns often come with an internal consistency, making them particularly suitable to be described by human language building on top of concepts~\cite{liu2020towards}. 

There is recently a surge of interest on this topic from both  academia and industry. HCOMP recently launched the CATS4ML challenge\footnote{\url{https://cats4ml.humancomputation.com}} to leverage crowdsourcing for unknown unknowns discovery; Facebook recently introduced the Dynabench platform\footnote{\url{https://dynabench.org}} for a similar purpose. 

Despite that, the recent effort has focused on data only, taking a bottom-up approach, that is: by collecting better data we hope the machines will learn what is needed. The assumption however comes without any theoretical guarantee or strong empirical evidence. 

\smallskip
\noindent\textbf{Hybrid Human-Machine Systems}  tackle the process of solving classification problems by leveraging both humans and machines ~\cite{Raghu2019,WilderHK20}. Initial work focused on very interesting ways to do this, from learning crowd vote aggregation models from ``features'' of the crowd task~\cite{Kamar_2012_combining}, to leveraging crowds to learn features of ML models, as in the brilliant paper by Bernstein and colleagues as well as  others~\cite{flock_2015,Carlos_pattern}.
More recently, proposals have emerged based on training an ML model for a task and then first using that model to classify, then asking humans if that model's confidence is not high enough \cite{Law_hearth_cscw18}. 
The effectiveness of such an approach is, consequently, heavily dependent on the reliability of machine confidence, which has shown to be very poor especially for deep learning \cite{Guo2017,Balda2020}.

\section{What Can Human Computation Do?}
Our literature analysis points to the fact that research from existing efforts only provides partial solutions to desirable ML systems. The problem of model rejection has been seldom discussed in the human computation community. 

The problem is closely related to the ML reliability issue that is heatedly discussed across many other communities, together with other issues such as transparency and fairness. Within Computer Science, discussions have been revolving around the relation between systems and people, e.g., the importance of human centrality. A visible trend is the fast growing work of human-AI interaction \cite{amershi2019guidelines,liao2020questioning}.
Much of those work takes the angle of humans as users or stakeholders; in comparison, the computational roles of humans in the process of better making ML systems or in the functioning of hybrid human-AI systems are seemingly less discussed. We note that human involvement in the system (creation) is key to bridge the gap between the need of stakeholders and the engineering of the system, hence of great scientific relevance to the engineering communities on ML, data, and systems. 



Human computation started with the very idea of leveraging human intelligence to solve tasks that are beyond the capability of automated systems, considering specifically the computational roles of humans without ignoring the personal and social properties. Responding to the model rejection problem, the key research question is the following:

\smallskip
\noindent\textbf{RQ:} \emph{How can human computation provide an approach to tell when machine learning systems fail?} 
Answers to this question can provide guidance for collecting high-utility data for model training and allow for safe decision delegation to machines. Having such an approach can, therefore, largely benefit data creation and hybrid decision-making, and together, promise a human-in-the-loop ML system that can be relied upon. 
We extrapolate a non-exhaustive list of sub-questions as follows:

\smallskip
\noindent
\textbf{SRQ 1:} \emph{What are the proper metrics for the cost of using ML with rejection?} 
Metrics should be re-considered to measure the cost-effectiveness of a hybrid human-AI system with a selective classifier whose prediction can be rejected. Following cost needs to be taken into account in the functioning of the system: 1) cost of wrong predictions by the ML model. Such cost is task-specific: the false positive and false negative should be weighted according to the task; 2) cost of human-made decisions. In the creation of the system, cost induced by human involvement in creating the classifier and the rejector should also be considered.

\smallskip
\noindent
\textbf{SRQ 2:} \emph{How to involve the right stakeholders to report on machine learning failures?} 
We can imagine that not all failures are easily detectable by random crowds, especially in social contexts where the perception of the quality of the services is dependent on personal preferences or cultural background. Opening a channel where stakeholders can effectively report on their experiences is the first key step to the rejection problem. The ``how'' in this question is, therefore, relevant to both the ``who'' and ``through which means''. 

\smallskip
\noindent\textbf{SRQ 3:} \emph{How to effectively characterize machine learning failures?} 
Characterization of failures can either be done on a per-item basis, i.e., using examples as description, or on the conceptual level. The latter would be preferred to provide a cognizable description of ``when the model fails'' to developers and stakeholders. 

Feasibility of such a ``symbolic'' approach is less a concern given the internal consistency of machine unknowns. In addition to the question of ``which form'' the description should be, it is also important to consider ``what materials'' to use in human characterization. We argue that the important factor is the involvement of models, through e.g., explainable AI techniques, such that the internal mechanism of the model can be exposed to allow for more effective identification of the failure reasons. In fact, recent work has shown that human computation can be a favorable approach to explanation itself~\cite{balayn2021you}.

\smallskip
\noindent\textbf{SRQ 4:} \emph{How to build the rejector?} A smart rejector can be built based on human feedback on machine failures. This can be done in a data-driven way like the normal ML models are trained, or through a hybrid data- and knowledge-driven method that allows for more explicit control over the items on which the prediction should be rejected. Reliability of human feedback should be considered, as to how human-labeled data has been used for ML training. 

In summary, we propose a new frame for evaluation where we argue that i) rejection - and related relevant metrics - should be a first class citizen of ML research, both theory and practice, that ii) hcomp is a promising way to go, but that iii) it requires very different methods than hcomp for data labeling.

\bibliography{main}

\end{document}